\documentclass[journal=jacsat,manuscript=article,layout=twocolumn]{achemso}

\usepackage{chemformula} 
\usepackage[T1]{fontenc} 

\usepackage{url}
\usepackage{float}
\usepackage{caption}
\usepackage{amsmath}
\usepackage{graphicx}
\usepackage{listings}
\usepackage{float}
\usepackage{tabularx}
\usepackage{amsfonts}
\usepackage{hyperref}
\usepackage{listings}
\usepackage{verbatim}
\usepackage{xcolor}
\usepackage{soul}
\usepackage{fancyvrb}
\usepackage{mdframed}
\usepackage{adjustbox}
\usepackage{makecell}
\usepackage{comment}

\definecolor{codebg}{RGB}{240,240,240}
\definecolor{codegreen}{RGB}{0,128,0}
\definecolor{codeblue}{RGB}{0,0,128}
\definecolor{codepurple}{RGB}{128,0,128}
\definecolor{codegray}{RGB}{128,128,128}

\lstdefinestyle{mystyle}{
    backgroundcolor=\color{codebg},
    commentstyle=\color{codegreen},
    keywordstyle=\color{codeblue},
    numberstyle=\tiny\color{codegray},
    stringstyle=\color{codepurple},
    basicstyle=\ttfamily\small, 
    breakatwhitespace=false,
    breaklines=true,
    captionpos=b,
    keepspaces=true,
    numbers=left,
    numbersep=5pt,
    showspaces=false,
    showstringspaces=false,
    showtabs=false,
    tabsize=2
}

\newcommand{\myverbatim}[1]{%
  \colorbox{codebg}{%
    \begin{minipage}{\dimexpr\linewidth-2\fboxsep}%
      \begin{verbatim}
      #1
      \end{verbatim}
    \end{minipage}%
  }%
}

\usepackage{caption}

\author{Albert Bou}
\affiliation[upf]
{Computational Science Laboratory, Universitat Pompeu Fabra, Barcelona Biomedical Research Park (PRBB), C Dr. Aiguader 88, 08003 Barcelona, Spain.}
\alsoaffiliation[acellera]{Acellera Labs, C Dr Trueta 183, 08005, Barcelona, Spain}

\author{Morgan Thomas}
\affiliation[upf]
{Computational Science Laboratory, Universitat Pompeu Fabra, Barcelona Biomedical Research Park (PRBB), C Dr. Aiguader 88, 08003 Barcelona, Spain.}

\author{Sebastian Dittert}
\affiliation[upf]
{Computational Science Laboratory, Universitat Pompeu Fabra, Barcelona Biomedical Research Park (PRBB), C Dr. Aiguader 88, 08003 Barcelona, Spain.}

\author{Carles Navarro}
\affiliation[acellera]{Acellera Labs, C Dr Trueta 183, 08005, Barcelona, Spain}

\author{Maciej Majewski}
\affiliation[acellera]{Acellera Labs, C Dr Trueta 183, 08005, Barcelona, Spain}

\author{Ye Wang}
\affiliation[BIOGEN]{Biogen Research and Development, 225 Binney Street, Cambridge, Massachusetts 02142, United States}

\author{Shivam Patel}
\affiliation[PSI]{Psivant Therapeutics, 451 D Street, Boston, Massachusetts 02210, United States}

\author{Gary Tresadern}
\affiliation[JN]{In Silico Discovery, Janssen Research \& Development, Janssen Pharmaceutica N. V., Turnhoutseweg 30, B-2340 Beerse, Belgium.}

\author{Mazen Ahmad}
\affiliation[JN]{In Silico Discovery, Janssen Research \& Development, Janssen Pharmaceutica N. V., Turnhoutseweg 30, B-2340 Beerse, Belgium.}

\author{Vincent Moens}
\affiliation{PyTorch team, Meta, 11 – 21 Canal Reach, London, N1C 4DB, UK}

\author{Woody Sherman}
\affiliation[PSI]{Psivant Therapeutics, 451 D Street, Boston, Massachusetts 02210, United States}

\author{Simone Sciabola}
\affiliation[BIOGEN]{Biogen Research and Development, 225 Binney Street, Cambridge, Massachusetts 02142, United States}

\author{Gianni De Fabritiis}
\alsoaffiliation[upf]
{Computational Science Laboratory, Universitat Pompeu Fabra, Barcelona Biomedical Research Park (PRBB), C Dr. Aiguader 88, 08003 Barcelona, Spain.}
\alsoaffiliation[acellera]{Acellera Labs, C Dr Trueta 183, 08005, Barcelona, Spain}
\affiliation[icrea]{Instituci\'o Catalana de Recerca i Estudis Avan\c{c}ats (ICREA), Passeig Lluis Companys 23, 08010 Barcelona, Spain}
\email{g.defabritiis@gmail.com}

\title[An \textsf{achemso} demo]
  {ACEGEN: 
 Reinforcement learning of generative chemical agents for drug discovery}

\begin{document}

\begin{tocentry}
\centering
\includegraphics[width=4.5cm,height=4.5cm]{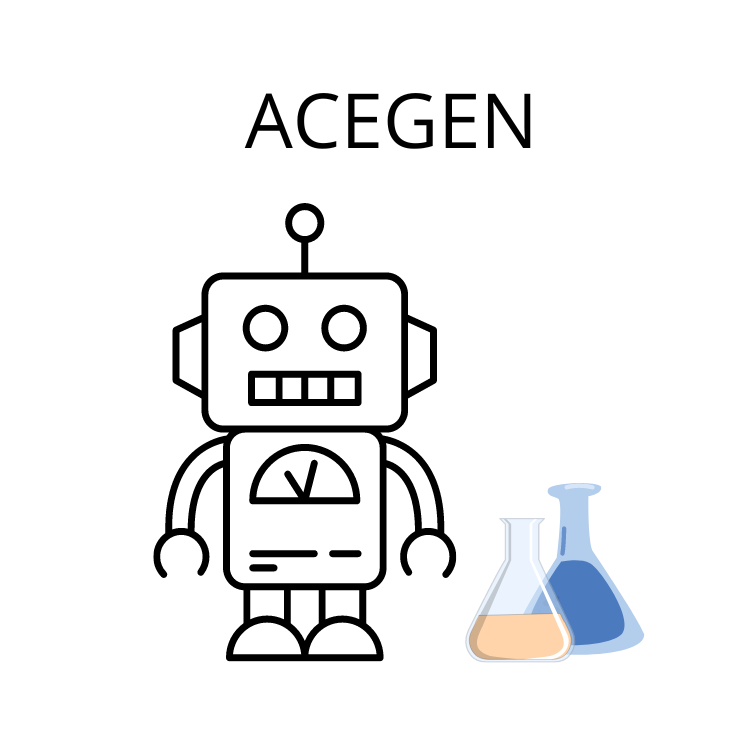}
\end{tocentry}

\begin{abstract}
In recent years, reinforcement learning (RL) has emerged as a valuable tool in drug design, offering the potential to propose and optimize molecules with desired properties. However, striking a balance between capabilities, flexibility, reliability, and efficiency remains challenging due to the complexity of advanced RL algorithms and the significant reliance on specialized code.
In this work, we introduce ACEGEN, a comprehensive and streamlined toolkit tailored for generative drug design, built using TorchRL, a modern RL library that offers thoroughly tested reusable components. We validate ACEGEN by benchmarking against other published generative modeling algorithms and show comparable or improved performance. We also show examples of ACEGEN applied in multiple drug discovery case studies.
ACEGEN is accessible at \url{https://github.com/acellera/acegen-open} and available for use under the MIT license.
\end{abstract}

\section{Introduction}

Drug design is a complex process that involves the identification of biomolecules that have an optimal balance of multiple properties, such as potency, selectivity, bioavailability, and toxicity.
In recent years, a diversity of generative modeling solutions have been proposed as a promising approach to partially automate the process of proposing new molecules that simultaneously improve multiple desired properties in the design-make-test-analysis cycle \citep{elton2019generativereview, wang2022generativereview}. These models typically employ machine learning algorithms to generate molecular candidates, but it remains challenging to efficiently search the vast chemical space to identify molecules with optimal properties \citep{brown2019guacamol}, which is so large that it is not practically enumerable in a naive manner.

Reinforcement learning (RL) has emerged as a possible solution to explore this chemical space \cite{olivecrona2017reinvent, popova2018reinforce} with an increasing focus on efficiency \cite{thomas2022augmented, guo2022improving}. RL \cite{sutton2018reinforcement} is a family of machine learning algorithms that use feedback as a learning signal to guide a decision-making process. Thus, RL algorithms can adapt a molecule building decision making progress to achieve certain characteristics of molecules resulting in novel molecules with desirable properties. This effectively constitutes a search strategy of chemical space. RL requires a reward function that assigns a value to a molecule, tailored specifically for the application or research question being addressed. The RL algorithm then seeks to maximize this value, therefore, adapting the navigation through the decision making progress and hence chemical space. This is particularly interesting when the available chemical space cannot be enumerated and filtered using the scoring function, either because the scoring functions are very slow or because the search space is very large.

There are some key differences between the application of RL classically and the application to search chemical space for drug design. Classically RL algorithms are developed and evaluated in game theory which usually has a well defined environment and a clear objective, for example, complete this level, or win the game. This is a huge discrepancy compared to drug discovery and design \cite{thomas2022applications}, where the objective is multi-faceted (an efficacious, non-toxic, and bioavailable molecule) and measured by proxies that poorly define the true objective e.g., optimizing the predicted binding affinity to a protein compared to the true objective of sufficiently perturbing a cellular pathway. This is unfortunately necessary due to the cost of running experimental assays. In light of this expected error between optimized proxy and true objective, it is important to propose diverse solutions \cite{renz2024diverse} to increase the chance of identifying a molecule of interest with all desirable properties for further development - another contrast to classical RL where a single successful solution is often sufficient. Despite these challenges, RL has shown promising preliminary results with successful application to drug design across a number of different molecular representations and model architectures \cite{du2024machine}.

However, currently available implementations for drug discovery using reinforcement learning (RL) rely heavily on custom code \cite{blaschke2020reinvent, gao2022sample}. This approach tends to favor redundancy, complexity, and limited efficiency, making it difficult to incorporate a diverse set of solutions. However, the RL community has already developed solutions to address these challenges. TorchRL \cite{bou2023torchrl}, a comprehensive RL library, offers well-tested, independent state-of-the-art RL components. In this work, we adopt TorchRL's components as building blocks to assemble efficient and reliable drug discovery agents—a practice already successfully applied in diverse domains such as drone control \cite{xu2024omnidrones} and combinatorial optimization \cite{berto2023rl4co}.
This approach also fosters algorithmic research by enabling the encapsulation of new ideas within new components, which can seamlessly integrate with existing ones.
TorchRL operates within the PyTorch ecosystem \cite{paszke2019pytorch}, ensuring not only high-quality standards but also maintenance over time and future development. 

\begin{figure*}
\centering
\includegraphics[width=\textwidth]{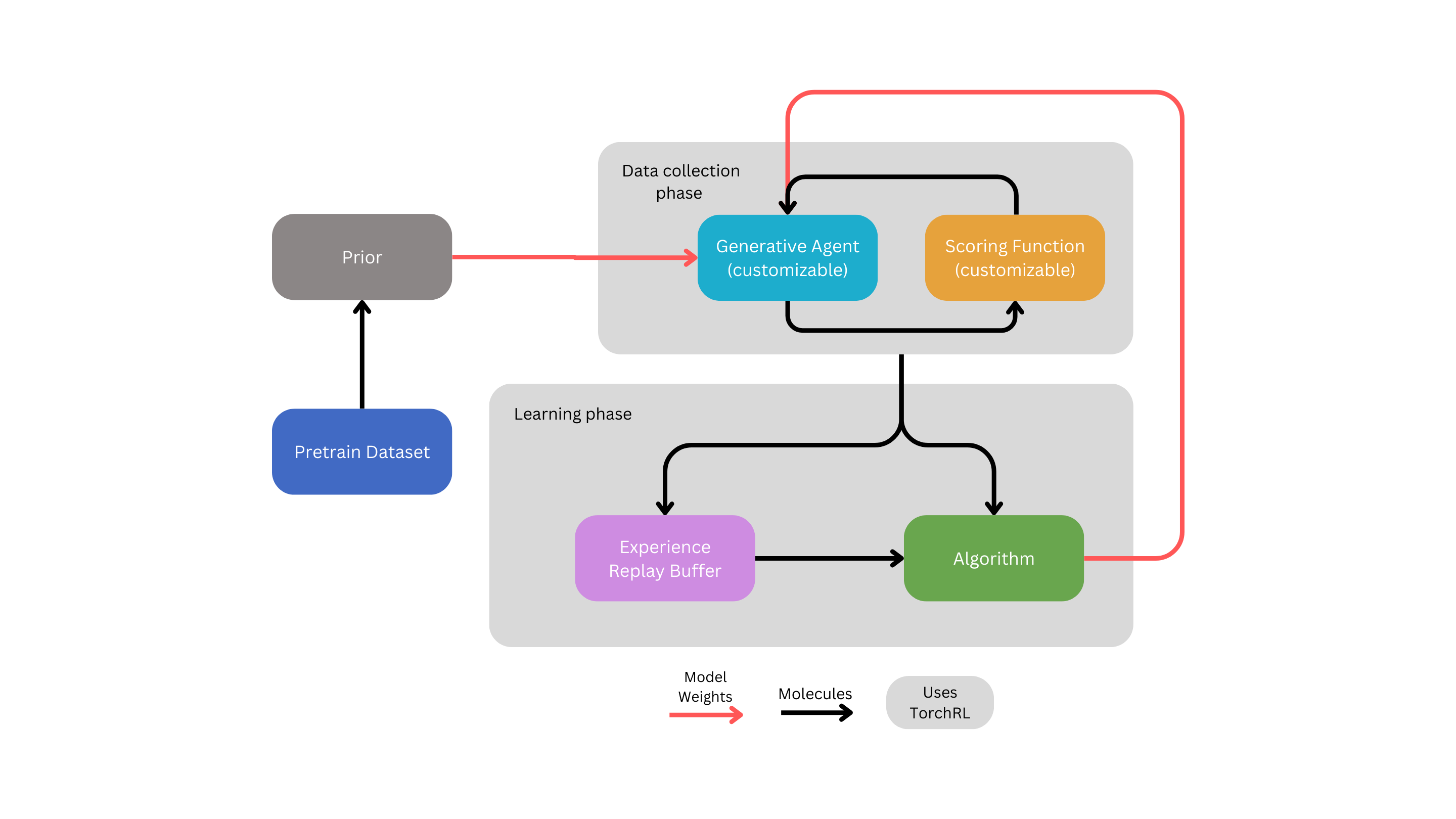}
\caption{General overview of any of the ACEGEN implementations. Different ACEGEN implementations vary in the algorithms used, and allow to customize the generative models and the scoring functions.}
\label{fig:workflow}
\end{figure*}

To showcase the advantages of ACEGEN, we implement and evaluate well-known language-based algorithms for drug design. An overview of the workflow for ACEGEN implementation is shown in \autoref{fig:workflow}. Language models can learn complex patterns in text and generate novel sequences, including molecular structures \cite{segler2018rnn, gomez2018automatic}.
Studies have focused on utilizing the Simplified Molecular Input Line Entry System (SMILES) \cite{weininger1988smiles}, a string notation system used to represent chemical structures in a compact and standardized manner. Moreover, several prior studies have demonstrated the benefits of combining generative language models and RL \cite{olivecrona2017reinvent, fialkova2021libinvent, guo2022improving}.
Therefore, we re-implement three REINFORCE-based algorithms: REINFORCE \cite{williams1992simple}, REINVENT \cite{olivecrona2017reinvent}, and AHC \cite{thomas2022augmented}. Additionally, we adapt general RL algorithms such as A2C \cite{mnih2016asynchronous} and PPO \cite{schulman2017proximal} to our problem setting. The REINVENT and AHC methods utilize some level of experience replay; therefore, we also incorporate experience replay into REINFORCE. Similarly, we test PPOD \cite{libardi2021guided}, a PPO-based algorithm adapted for experience replay.

To demonstrate the possible use cases and applications of ACEGEN,  we have benchmarked these RL algorithms on a sample efficiency benchmark while considering chemistry, conducted an ablation study to better understand the components of a specific RL algorithm, tested a more challenging drug design relevant objective, and lastly presented RL for molecular constrained generative design.
ACEGEN code is accessible open source under the MIT license at \url{https://github.com/acellera/acegen-open} to facilitate uptake by the community.

\section{Methods}
 
\subsection{Reinforcement learning setting}

Reinforcement learning (RL) tasks are formalized as Markov Decision Processes (MDPs) \cite{sutton2018reinforcement}, described by the quintuple $\langle S, A, R, P, \rho_0 \rangle$. 
Here, $S$ is the set of all possible states in the problem space, $A$ is the set of valid actions available to the agent, and $R: S \times A \times S \to \mathbb{R}$ is the reward function, which assigns numerical value to the transition from one state to the next given the action taken. The function $P: S \times A \to \mathcal{P}(S)$ is the transition probability, where $P(s_{t+1}|s,a)$ is the probability of transitioning to state $s_{t+1}$ from the current state $s$ under action $a$. Lastly, $\rho_0$ signifies the initial state distribution.

This can be applied to the problem of designing molecules sequentially. In this context, a parameterized RL policy $\pi_{\theta}$ can navigate molecule design by selecting actions (molecular edits) $a_t$ at a given state $s_t$ (partially built molecule) until molecule building is terminated. The reward function assigns a scalar value that captures desirability, either at each step or upon termination of molecule building. Where $\pi_{\theta}(a_t |s_t)$ denotes the probability of taking action $a_t$ in state $s_t$ from the policy function parameterized by $\theta$. Meanwhile, $\tau$ represents the episode, or sequence of actions needed to construct a molecule. Thus, $P(\tau|\theta)$ is the probability of the full trajectory $\tau$ given the policy parameters $\theta$, and $R(\tau)$ is the cumulative sum of rewards over the trajectory $\tau$.

The goal of policy-based RL algorithms is to optimize the parameters of the policy $\pi_{\theta}$ to maximize $R(\tau)$. Different methods within the family of policy gradient methods \cite{sutton2018reinforcement}, such as REINFORCE \cite{williams1992simple}, A2C (Advantage Actor-Critic) \cite{mnih2016asynchronous}, and PPO (Proximal Policy Optimization) \cite{schulman2017proximal}, are commonly used for this task. 

Additionally, various techniques can aid in training. Reward shaping modifies the reward function to provide additional feedback to the agent during training, potentially providing more informative signals. Experience replay stores past experiences in a buffer and randomly samples from it during training, which can improve sample efficiency and stabilize learning. Incorporating penalty terms, such as a Kullback-Leibler (KL) divergence loss term, can encourage the agent to stay close to a reference or prior policy, helping to maintain stability. Finally, ranking and selecting only the best K molecules in each given batch can improve efficiency. ACEGEN integrates all the aforementioned algorithms and additional techniques to form a comprehensive suite of state-of-the-art methods for molecular generation.

\subsection{Chemical language generative models}

Molecular string representations \cite{wiswesser1985historic} convert molecular graphs into strings and vice versa. This representation therefore formulates the task of molecular generation as a natural language processing (NLP) \cite{ozturk2020exploring} problem, and the policy models used to sequentially generate molecules represented as strings are called Chemical Language Models (CLMs) \cite{grisoni2023chemical}. In this context, the resulting action space is discrete, with each action $a$ denoted as a token, and each non-terminal state $s_t$ a partially complete SMILES string. Each episode begins with a single special start token, e.g., "GO" and can last for a varying number of steps. The episode ends when the agent chooses another special token called the stop token e.g., "EOS".

ACEGEN currently provides an environment for language model experimentation including SMILES \cite{weininger1988smiles}, DeepSMILES \cite{o2023deepsmiles}, SELFIES \cite{krenn2020selfies}, AtomInSmiles \cite{ucak2023ais} and SAFE \cite{noutahi2024gotta} grammars, complemented by a user-friendly vocabulary class. Section A in the supporting information illustrates how the vocabulary and environment can be easily created and utilized for data generation in ACEGEN.

CLMs are first trained unsupervised on a bulk of unlabeled data, aimed at learning to generate valid molecules, achieved through the application of the teacher enforcing method. \cite{lamb2016professor}. The pre-trained CLM is the starting policy function $\pi_\theta$ that can be further trained to optimize a specific objective with RL. Moreover, this prior policy can be used as an anchor point from which the new RL policy should not deviate excessively.

ACEGEN currently provides pre-trained models for several architectures. We provide GRU  \cite{gru} policies pre-trained on two datasets: ChEMBL \cite{gaulton2012chembl} and ZINC \cite{gomez2018automatic}. We also provide an LSTM \cite{lstm} policy pre-trained on ChEMBL \cite{gaulton2012chembl},  and a GPT2 \cite{radford2019language} policy pre-trained on the Enamine Database REAL lead-like compounds \cite{shivanyuk2007enamine}.  Our repository  provides ready-to-use architectures for LSTM \cite{lstm}, GRU \cite{gru}, GPT2 \cite{radford2019language}
, Llama2 \cite{touvron2023llama} and Mamba \cite{gu2023mamba}
policies. Moreover, users can include any other architectures of their choice following our step-by-step tutorial that explain how to do the integration, which is possible without modifying any ACEGEN internals.

Finally, ACEGEN also provides pre-training script to train language models, either from ACEGEN or custom. Our code is engineered to adapt to the available computational resources, whether a single GPU or a distributed setup spanning multiple machines and GPUs. This adaptability allows to efficiently train models on datasets of large size. 

\subsection{Scoring and Evaluation of Molecules}
\label{sec:molscore}

Scoring functions must reflect real-world drug design scenarios and be flexible enough to apply to a range of drug design challenges. Often, overly simplistic objective functions are used for optimization, such as penalized logP \cite{jin2018junction}, or complex solutions tailored to one particular generative model \cite{blaschke2020reinvent}. 
ACEGEN allows the integration of custom scoring functions by providing the flexibility to define them as Python methods that accept strings and return numerical values. 

Section B in the supporting information showcases how scoring functions can be implemented easily in ACEGEN. Additionally, the library offers a detailed tutorial, guiding users through the process of implementing and incorporating custom scoring functions seamlessly into the entire workflow to train agents with them.

In this paper for benchmarking we have used the MolScore \cite{thomas2024molscore} library, which offers a broad range of drug design relevant scoring functions, diversity filters, support for curriculum learning and also contains a benchmarking mode including MolOpt \cite{gao2022sample}, GuacaMol \cite{brown2019guacamol}, and others. 

\subsection{RL Agents Training}

ACEGEN provides training for RL agents utilizing the following methods: REINFORCE \cite{williams1992simple}, REINVENT \cite{blaschke2020reinvent}, AHC \cite{thomas2022augmented}, A2C \cite{mnih2016asynchronous}, and PPO \cite{schulman2017proximal}, as well as an adapted version of the PPOD algorithm \cite{libardi2021guided}. We deviate from the exact implementation of PPOD by omitting the use of an initial expert demonstration, replaying only episodes with high rewards (instead of episodes with high-value predictions), and employing a fixed amount of replay data per batch. All methods are fully configurable, allowing the use of custom scoring functions and models beyond those already provided by the library.

REINVENT and AHC can be considered extensions of the REINFORCE algorithm. Specifically, REINVENT incorporates reward shaping, experience replay, and a penalty for high-likelihood sequences into the basic REINFORCE framework. Additionally, AHC ranks and selects the top K molecules in each batch of data.
We find that using the same reward shaping is not compatible with advantage-based methods, and the high-likelihood penalty term is detrimental. Therefore, for A2C, PPO, and PPOD, we introduce a term to the loss function based on the Kullback-Leibler divergence (KL) between the actor policy and a prior policy. This term serves to penalize the policy for deviating too much from a prior, similar in concept to reward shaping in REINVENT and AHC.
Incorporating KL constraints is a common practice in research papers aligning language models with custom reward functions, as seen in examples utilizing human feedback \cite{menick2022teaching, ouyang2022training, bai2022training}.

\subsection{De-novo, decorative and fragment-linking generation}
\label{sec:promptsmiles}

In drug discovery pipelines, generating molecules from scratch may not always suffice. To meet diverse requirements, ACEGEN offers multiple sampling modes. PromptSMILES \cite{thomas2024promptsmiles} is a simple method enabling constrained molecule generation using models pre-trained solely with teacher enforcement on full SMILES, making it possible to generate molecules while adhering to specific chemical sub-structures. Specifically, in addition to de-novo generation, ACEGEN scripts allow for easy configuration of scaffold decoration and fragment-linking molecule generation modes. Constrained sampling modifies only the data collection behavior, operating independently of all other agent components, in line with the TorchRL philosophy. Tutorials on performing constrained generation are provided within the repository.

\section{Results}

To showcase some of the use cases of ACEGEN, we have benchmarked multiple methods in chemistry benchmarks, conducted an ablation study on REINVENT, tested a challenging drug design objective, and explored constrained RL with PromptSMILES.

\subsection{Benchmarking RL performance}

To assess different RL algorithms, we have compared performance on the Practical Molecular Optimization (MolOpt) benchmark \cite{gao2022sample} as implemented in MolScore \cite{thomas2024molscore}. This benchmark encompasses 23 distinct tasks associated with different objectives to be optimized within a budget of 10,000 molecules. All algorithms use the same GRU policy model architecture as in the original MolOpt benchmark, implemented in ACEGEN and pretrained on a curated subset of ChEMBL \cite{gaulton2012chembl}. We compared REINFORCE \cite{williams1992simple}, REINVENT with the hyperparameters from the original paper \cite{blaschke2020reinvent}, REINVENT-MolOpt with optimized hyperparameters for the MolOpt benchmark \cite{gao2022sample}, and AHC with the hyperparameters from the original paper \cite{thomas2022augmented}. Note that we utilize experience replay in all of the aforementioned algorithms
Additionally, we also test A2C \cite{mnih2016asynchronous}, PPO \cite{schulman2017proximal} which does not utilize any experience replay, and PPOD \cite{libardi2021guided} which is an adaptation of PPO to accommodate experience replay. All algorithms are ACEGEN implementations, and the hyperparameter values used for all of them are provided in our repository.  As a sanity check, we compared our implementation of REINVENT with the implementation used in the MolOpt paper. Our simplified re-implementation achieved better results with faster training. These results are shown in section C in the supporting information.

\begin{table*}[ht!]
\caption{Algorithm comparison for the Area Under the Curve (AUC) of the top 10 molecules on MolOpt benchmark objectives. This metric captures the sample efficiency in identifying 10 desirable molecules with respect to the objective. Each algorithm was run 5 times with different seeds, and results were averaged.}
\centering
\small 
\resizebox{\linewidth}{!}{
\begin{tabular}{|l|c|c|c|c|c|c|c|}\hline
 & ACEGEN & ACEGEN & ACEGEN & ACEGEN & ACEGEN & ACEGEN & ACEGEN \\
Task & REINFORCE & REINVENT & REINVENT-MolOpt & AHC & A2C & PPO & PPOD \\ \hline
Albuterol\_similarity & 0.68 $\pm$ 0.03 & 0.69 $\pm$ 0.02 & 0.90 $\pm$ 0.01 & 0.77 $\pm$ 0.02 & 0.82 $\pm$ 0.04 & 0.93 $\pm$ 0.02 & \textbf{0.94 $\pm$ 0.00} \\
Amlodipine\_MPO & 0.55 $\pm$ 0.01 & 0.56 $\pm$ 0.01 & 0.65 $\pm$ 0.06 & 0.56 $\pm$ 0.01 & 0.55 $\pm$ 0.01 & 0.58 $\pm$ 0.03 & \textbf{0.68 $\pm$ 0.02} \\
C7H8N2O2 & 0.83 $\pm$ 0.01 & 0.82 $\pm$ 0.03 & \textbf{0.90 $\pm$ 0.02} & 0.76 $\pm$ 0.04 & 0.84 $\pm$ 0.04 & 0.89 $\pm$ 0.01 & 0.89 $\pm$ 0.03 \\
C9H10N2O2PF2Cl & 0.70 $\pm$ 0.02 & 0.70 $\pm$ 0.02 & 0.76 $\pm$ 0.03 & 0.68 $\pm$ 0.02 & 0.69 $\pm$ 0.03 & 0.66 $\pm$ 0.02 & \textbf{0.79 $\pm$ 0.02} \\
Celecoxxib\_rediscovery & 0.63 $\pm$ 0.02 & 0.64 $\pm$ 0.03 & 0.77 $\pm$ 0.02 & 0.72 $\pm$ 0.02 & 0.73 $\pm$ 0.06 & 0.65 $\pm$ 0.12 & \textbf{0.82 $\pm$ 0.03} \\
DRD2 & 0.98 $\pm$ 0.00 & 0.97 $\pm$ 0.00 & \textbf{0.99 $\pm$ 0.00} & 0.98 $\pm$ 0.01 & 0.98 $\pm$ 0.01 & \textbf{0.99 $\pm$ 0.00} & \textbf{0.99 $\pm$ 0.00} \\
Deco\_hop & 0.63 $\pm$ 0.00 & 0.63 $\pm$ 0.01 & \textbf{0.67 $\pm$ 0.01} & 0.64 $\pm$ 0.01 & 0.62 $\pm$ 0.00 & 0.62 $\pm$ 0.01 & 0.66 $\pm$ 0.02 \\
Fexofenadine\_MPO & 0.71 $\pm$ 0.01 & 0.71 $\pm$ 0.00 & \textbf{0.80 $\pm$ 0.03} & 0.72 $\pm$ 0.00 & 0.71 $\pm$ 0.00 & 0.73 $\pm$ 0.00 & 0.78 $\pm$ 0.01 \\
GSK3B & 0.84 $\pm$ 0.01 & 0.84 $\pm$ 0.02 & \textbf{0.92 $\pm$ 0.02} & 0.82 $\pm$ 0.01 & 0.85 $\pm$ 0.02 & 0.90 $\pm$ 0.02 & \textbf{0.92 $\pm$ 0.02} \\
JNK3 & 0.75 $\pm$ 0.03 & 0.75 $\pm$ 0.02 & 0.85 $\pm$ 0.04 & 0.75 $\pm$ 0.01 & 0.74 $\pm$ 0.06 & 0.80 $\pm$ 0.04 & \textbf{0.87 $\pm$ 0.02} \\
Median\_molecules\_1 & 0.26 $\pm$ 0.00 & 0.24 $\pm$ 0.00 & \textbf{0.36 $\pm$ 0.02} & 0.24 $\pm$ 0.00 & 0.31 $\pm$ 0.01 & 0.33 $\pm$ 0.02 & 0.35 $\pm$ 0.02 \\
Median\_molecules\_2 & 0.22 $\pm$ 0.00 & 0.22 $\pm$ 0.00 & 0.28 $\pm$ 0.01 & 0.24 $\pm$ 0.00 & 0.25 $\pm$ 0.01 & 0.25 $\pm$ 0.02 & \textbf{0.29 $\pm$ 0.01} \\
Mestranol\_similarity & 0.60 $\pm$ 0.03 & 0.55 $\pm$ 0.04 & 0.85 $\pm$ 0.07 & 0.66 $\pm$ 0.04 & 0.69 $\pm$ 0.07 & 0.75 $\pm$ 0.15 & \textbf{0.89 $\pm$ 0.05} \\
Osimertinib\_MPO & 0.82 $\pm$ 0.01 & 0.82 $\pm$ 0.00 & \textbf{0.86 $\pm$ 0.01} & 0.83 $\pm$ 0.00 & 0.81 $\pm$ 0.01 & 0.82 $\pm$ 0.01 & 0.84 $\pm$ 0.00 \\
Perindopril\_MPO & 0.48 $\pm$ 0.01 & 0.47 $\pm$ 0.00 & \textbf{0.54 $\pm$ 0.01} & 0.47 $\pm$ 0.01 & 0.48 $\pm$ 0.00 & 0.50 $\pm$ 0.01 & 0.53 $\pm$ 0.00 \\
QED & \textbf{0.94 $\pm$ 0.00} & \textbf{0.94 $\pm$ 0.00} & \textbf{0.94 $\pm$ 0.00} & \textbf{0.94 $\pm$ 0.00} & \textbf{0.94 $\pm$ 0.00} & \textbf{0.94 $\pm$ 0.00} & \textbf{0.94 $\pm$ 0.00} \\
Ranolazine\_MPO & 0.70 $\pm$ 0.01 & 0.69 $\pm$ 0.00 & \textbf{0.76 $\pm$ 0.01} & 0.70 $\pm$ 0.00 & 0.74 $\pm$ 0.01 & 0.73 $\pm$ 0.01 & 0.75 $\pm$ 0.00 \\
Scaffold\_hop & 0.80 $\pm$ 0.00 & 0.79 $\pm$ 0.00 & \textbf{0.86 $\pm$ 0.02} & 0.80 $\pm$ 0.01 & 0.80 $\pm$ 0.00 & 0.80 $\pm$ 0.02 & 0.84 $\pm$ 0.03 \\
Sitagliptin\_MPO & 0.34 $\pm$ 0.02 & 0.33 $\pm$ 0.01 & 0.38 $\pm$ 0.03 & 0.33 $\pm$ 0.02 & \textbf{0.39 $\pm$ 0.02} & 0.32 $\pm$ 0.02 & \textbf{0.39 $\pm$ 0.02} \\
Thiothixene\_rediscovery & 0.41 $\pm$ 0.01 & 0.41 $\pm$ 0.00 & 0.56 $\pm$ 0.04 & 0.45 $\pm$ 0.02 & 0.48 $\pm$ 0.04 & 0.48 $\pm$ 0.06 & \textbf{0.58 $\pm$ 0.09} \\
Troglitazone\_rediscovery & 0.31 $\pm$ 0.02 & 0.31 $\pm$ 0.02 & 0.47 $\pm$ 0.05 & 0.34 $\pm$ 0.01 & 0.35 $\pm$ 0.02 & 0.46 $\pm$ 0.07 & \textbf{0.52 $\pm$ 0.06} \\
Valsartan\_smarts & \textbf{0.03 $\pm$ 0.00} & 0.02 $\pm$ 0.00 & 0.02 $\pm$ 0.00 & 0.02 $\pm$ 0.00 & 0.02 $\pm$ 0.00 & \textbf{0.03 $\pm$ 0.00} & \textbf{0.03 $\pm$ 0.00} \\
Zaleplon\_MPO & 0.47 $\pm$ 0.01 & 0.47 $\pm$ 0.01 & \textbf{0.52 $\pm$ 0.01} & 0.48 $\pm$ 0.01 & 0.47 $\pm$ 0.01 & 0.50 $\pm$ 0.02 & \textbf{0.52 $\pm$ 0.01} \\ \hline
Total & 13.67 & 13.60 & 15.65 & 13.91 & 14.27 & 14.65 & \textbf{15.80} \\ \hline
\end{tabular}
}
\label{tab:auc10}
\end{table*}

\begin{figure*}
\centering
\includegraphics[width=0.9\textwidth]{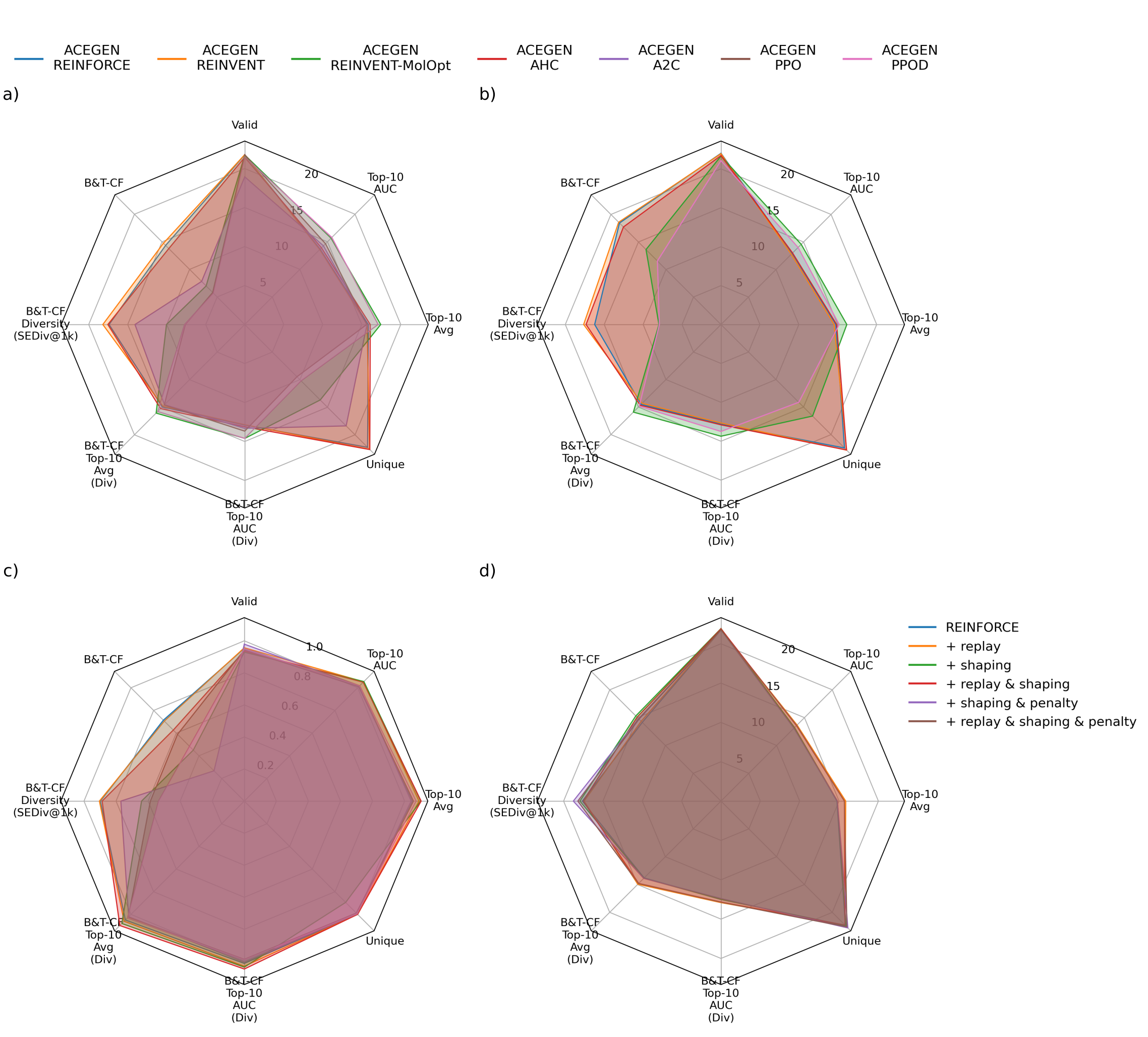}
\caption{Comparison of RL algorithms by radar plot visualization of metric performance for (a) the MolOpt benchmark reported in \autoref{tab:molopt_selectmetrics}, for (b) the MolOpt benchmark with chemistry requirements explicitly in the reward signal as reported in \autoref{tab:molopt-CF}, (c) the 5-HT$_{2A}$ case study as reported in \autoref{tab:5HT2A_avg_metrics}, and (d) the REINFORCE ablation study as reported in \autoref{tab:molopt_ablation}. The legend at the top of the figure applies to sub-plots (a), (b), and (c). Subplot (d) has the legend beside.}\label{fig:radars}
\end{figure*}

As proposed by the MolOpt benchmark authors, Table \autoref{tab:auc10} shows algorithm performance on the AUC of the top 10 molecules as an indication of the sample efficiency of identifying 10 desirable molecules with respect to the objective, a representative figure that might be carried forward to later stages of drug design. This shows that PPOD is state-of-the-art with respect to sample efficiency, followed by REINVENT-MolOpt and then PPO. To achieve a higher level understanding of performance, Table \ref{tab:molopt_selectmetrics} and \autoref{fig:radars}a shows a selection of metrics aimed to measure maximum performance, efficiency, and exploration (a common trade-off with exploitation and hence efficiency in RL). These results show that REINVENT-MolOpt achieves maximum performance as measured by the average top 10 molecules, PPOD maximum efficiency as measured by AUC of the top 10 molecules, and AHC maximum exploration as measured by the number of unique compounds generated.

\begin{table*}
\caption{Algorithm comparison on a selection of metrics analyzing the performance on the MolOpt benchmark. Valid is a sanity check that measures the proportion of valid molecules generated. Top-10 AUC is a measure of sample efficiency as shown in \ref{tab:auc10}. Top-10 Avg is a measure of the absolute best 10 molecules achieved with the budget. Unique is a proxy for exploration which measures the proportion of unique molecules generated. Basic and target chemistry filters (B\&T-CF) are the proportion of molecules after filtering out highly idiosyncratic molecules in general or with respect to the pretraining dataset. The B\&T-CF Top-10 AUC and B\&T-CF Top-10 Avg are recalculated on this subset while also enforcing that the 10 molecules identified offer diverse solutions (Div). Lastly, B\&T-CF Diversity is a proxy measure of exploration via sphere exclusion diversity \cite{thomas2021comparison} on this subset. Each algorithm was run 5 times with different seeds, the average score and variance was then summed over the 23 tasks to report the final score and standard deviation of the summed variance. A perfect score for all metrics is 23.}
\centering
\small 
\resizebox{\linewidth}{!}{
\centering
\begin{tabular}{|l|c|c|c|c|c|c|c|}
\hline
 & ACEGEN & ACEGEN & ACEGEN & ACEGEN & ACEGEN & ACEGEN & ACEGEN \\
Metric & REINFORCE & REINVENT & REINVENT-MolOpt & AHC & A2C & PPO & PPOD \\ \hline
Valid & 21.77 $\pm$ 0.03 & \textbf{21.78 $\pm$ 0.02} & 21.74 $\pm$ 0.05 & 21.45 $\pm$ 0.02 & 18.93 $\pm$ 0.51 & \textbf{21.78 $\pm$ 0.09} & 21.52 $\pm$ 0.15 \\
Top-10 Avg & 15.85 $\pm$ 0.10 & 15.69 $\pm$ 0.11 & \textbf{17.43 $\pm$ 0.17} & 16.09 $\pm$ 0.10 & 15.67 $\pm$ 0.22 & 15.63 $\pm$ 0.27 & 17.10 $\pm$ 0.19\\
Top-10 AUC & 13.67 $\pm$ 0.07 & 13.60 $\pm$ 0.08 & 15.65 $\pm$ 0.14 & 13.91 $\pm$ 0.08 & 14.27 $\pm$ 0.14 & 14.65 $\pm$ 0.23 & \textbf{15.80 $\pm$ 0.14} \\
Unique & 22.28 $\pm$ 0.10 & 22.63 $\pm$ 0.06 & 13.68 $\pm$ 0.31 & \textbf{22.68 $\pm$ 0.07} & 18.38 $\pm$ 0.50 & 9.47 $\pm$ 0.33 & 10.21 $\pm$ 0.35 \\
\hline
B\&T-CF & 14.34 $\pm$ 0.18 & \textbf{14.74 $\pm$ 0.15} & 7.00 $\pm$ 0.27 & 13.77 $\pm$ 0.12 & 7.82 $\pm$ 0.37 & 5.81 $\pm$ 0.26 & 5.71 $\pm$ 0.22 \\
B\&T-CF Top-10 Avg (Div) & 14.95 $\pm$ 0.12 & 14.83 $\pm$ 0.13  & \textbf{16.06 $\pm$ 0.16} & 15.25 $\pm$ 0.10 & 14.54 $\pm$ 0.18 & 14.60 $\pm$ 0.23 & 15.78 $\pm$ 0.15 \\
B\&T-CF Top-10 AUC (Div) & 12.91 $\pm$ 0.06 & 12.85 $\pm$ 0.08 & \textbf{14.61 $\pm$ 0.13} & 13.11 $\pm$ 0.08 & 13.32 $\pm$ 0.12 & 13.67 $\pm$ 0.17 & 14.60 $\pm$ 0.14 \\
B\&T-CF Diversity (SEDiv@1k) & 17.39 $\pm$ 0.15 & \textbf{18.19 $\pm$ 0.12} & 10.10 $\pm$ 0.31 & 17.55 $\pm$ 0.10 & 14.07 $\pm$ 0.47 & 7.61 $\pm$ 0.49 & 7.71 $\pm$ 0.39 \\
\hline
\end{tabular}
}

\label{tab:molopt_selectmetrics}
\end{table*}

\subsection{Benchmarking RL performance for practical drug discovery}

RL optimizes the policy for the cumulative future rewards as provided by either one scoring function or a sum of multiple scoring functions. However, in practical drug discovery, it is not straightforward to write the precise scoring functions that are needed and how to weigh them into a single scalar.  For example, scoring functions can sometimes not produce the desired chemistry, or show clear exploitation loopholes \cite{renz2019failure, langevin2022explaining}. In this setting, a sufficient element of regularization to a prior policy is useful, as the prior policy has learned a chemical space distribution based on a specified training dataset of choice which acts as a representation of desirable chemistry by examples. It is for this reason  that methods like REINVENT have been designed heuristically to contain terms that try to enforce regularization instead of entirely using the reward signal.

To account for this, we make a modification to the average and AUC of the top 10 molecules to ensure they are diverse by an ECFP4 Tanimoto similarity of less than 0.35 to each other. Moreover, we use sphere exclusion diversity (SEDiv@1k) of \textit{de novo} molecules \cite{thomas2021comparison} which measures the proportion of molecules needed to describe chemical space in a random sample of 1,000 molecules, as a proxy for exploration. Note this strategy of measuring diverse hits is of increasing interest \cite{xie2021much, renz2024diverse}. This is more representative than the number of unique molecules that could all reside in a close area of chemical space. Lastly, we remove \textit{de novo} molecules that do not pass a series of filters. This includes a basic chemistry filter (B-CF): logP less than or equal to 4.5, rotatable bond count less than or equal to 7, molecular weight in the range 150 to 650 Da, only contain atoms belonging to the following set $A\in\{C, S, O, N, H, F, Cl, Br\}$, and do not violate the substructure alerts as described in \cite{polykovskiy2020molecular}. Moreover on the assumption that the training dataset describes desirable chemistry by examples, a target chemistry filter (T-CF): logP and molecular weight within $\mu \pm 4 \sigma$ of the training dataset distribution, as well as, removing any molecule that comprises $>10\%$ novel atomic environment bits with respect to the reference molecules, as measured by ECFP4 bits 
(an example of this is shown in Figure S6)
. The combination of the basic and target chemistry filter is denoted (B\&T-CF).

Re-evaluating RL performance with these new metrics (\autoref{tab:molopt_selectmetrics}) it can be seen that a smaller proportion of molecules pass the chemistry filters for REINVENT-MolOpt, A2C, PPO, and PPOD, as expected due to decreased regularization terms to optimize efficiency. Despite this, REINVENT-MolOpt achieves the highest score for the average and AUC of the top 10 diverse molecules followed by PPOD. As a proxy for exploration, REINVENT contains the highest sphere exclusion diversity followed by AHC. Interestingly, REINFORCE maintains a high proportion of molecules passing the chemistry filters despite no explicit regularization term, outperforming REINVENT in absolute performance and efficiency. \autoref{tab:molopt-CF} and \autoref{fig:radars}b shows the alternative approach of explicitly including these requirements in the reward signal by the addition of the chemistry filters and a diversity filter to each objective. As expected, including these requirements explicitly in the reward function increases the proportion of molecules passing these filters. However, every algorithm drops in performance as measured by the average and AUC top 10 diverse molecules that pass the filters. In this case, this shows that better overall score optimization and efficiency (including those that measure the quality of chemistry) is achieved by implicit regularization to the prior policy. This could be a result of the specific implementation of the chemistry filters in the reward resulting in a score of 0 if they do not pass; however, optimal implementation of reward signal can in itself be considered an art. Therefore, simpler reward signals are usually better, and, achieving desirable chemistry by regularization enables the design of a simpler reward signal. The fine-tuning of the optimization algorithm, reward signals and regularization might be target-dependent and goes beyond the scope of this work, but it is enabled by ACEGEN.

\begin{table*}[h]
\centering
\resizebox{\linewidth}{!}{
\begin{tabular}{|l|ccccc|ccccc|}
\hline
 & \multicolumn{5}{c|}{MolOpt} & \multicolumn{5}{c|}{MolOpt-CF} \\
 \hline
 & ACEGEN & ACEGEN & ACEGEN & ACEGEN & ACEGEN & ACEGEN & ACEGEN & ACEGEN & ACEGEN & ACEGEN \\
Metric & REINFORCE & REINVENT & REINVENT-MolOpt & AHC & PPOD & REINFORCE & REINVENT & REINVENT-MolOpt & AHC & PPOD \\ \hline
Valid & 21.77 $\pm$ 0.03 & 21.78  $\pm$  0.02 & 21.74  $\pm$  0.05 & 21.45  $\pm$  0.02 & 21.52  $\pm$  0.15 & \textbf{22.00  $\pm$  0.02} & 21.95  $\pm$  0.02 & 21.65  $\pm$  0.06 & 21.71  $\pm$  0.03 & 21.05  $\pm$  0.14 \\

Top-10 Avg & 15.85 $\pm$ 0.10 & 15.69 $\pm$ 0.11 & \textbf{17.43 $\pm$ 0.17} & 16.09 $\pm$ 0.10 & 17.10 $\pm$ 0.19 & 14.76 $\pm$ 0.09 & 14.54 $\pm$ 0.10 & 16.17 $\pm$ 0.16 & 14.86 $\pm$ 0.08 & 15.18 $\pm$ 0.24 \\

Top-10 AUC & 13.67 $\pm$ 0.07 & 13.60 $\pm$ 0.08 & 15.65 $\pm$ 0.14 & 13.91 $\pm$ 0.08 & \textbf{15.80 $\pm$ 0.14} & 12.94 $\pm$ 0.07 & 12.78 $\pm$ 0.08 & 14.57 $\pm$ 0.14 & 13.00 $\pm$ 0.06 & 14.01 $\pm$ 0.18 \\

Unique & 22.28 $\pm$ 0.10 & 22.63 $\pm$ 0.06 & 13.68 $\pm$ 0.31 & 22.68 $\pm$ 0.07 & 10.21 $\pm$ 0.35 & 22.44 $\pm$ 0.05 & \textbf{22.83 $\pm$ 0.01} & 16.63 $\pm$ 0.20 & 22.80 $\pm$ 0.01 & 14.13 $\pm$ 0.36 \\
\hline
B\&T-CF & 14.34 $\pm$ 0.18 & 14.74 $\pm$ 0.15 & 7.00 $\pm$ 0.27 & 13.77 $\pm$ 0.12 & 5.71 $\pm$ 0.22 & 18.44 $\pm$ 0.06 & \textbf{18.59 $\pm$ 0.04} & 13.62 $\pm$ 0.16 & 17.72 $\pm$ 0.04 & 11.55 $\pm$ 0.30 \\

B\&T-CF Top-10 Avg (Div) & 14.95 $\pm$ 0.12 & 14.83 $\pm$ 0.13 & \textbf{16.06 $\pm$ 0.16} & 15.25 $\pm$ 0.10 & 15.78 $\pm$ 0.15 & 14.51 $\pm$ 0.09 & 14.33 $\pm$ 0.08 & 15.92 $\pm$ 0.13 & 14.68 $\pm$ 0.08 & 14.86 $\pm$ 0.22 \\

B\&T-CF Top-10 AUC (Div) & 12.91 $\pm$ 0.06 & 12.85 $\pm$ 0.08 & \textbf{14.61 $\pm$ 0.13} & 13.11 $\pm$ 0.08 & 14.60 $\pm$ 0.14 & 12.79 $\pm$ 0.06 & 12.66 $\pm$ 0.07 & 14.35 $\pm$ 0.12 & 12.89 $\pm$ 0.07 & 13.74 $\pm$ 0.16 \\

B\&T-CF Diversity (SEDiv@1k) & 17.39 $\pm$ 0.15 & \textbf{18.19 $\pm$ 0.12} & 10.03 $\pm$ 0.31 & 17.55 $\pm$ 0.10 & 7.71 $\pm$ 0.39 & 16.23 $\pm$ 0.13 & 17.65 $\pm$ 0.10 & 7.97 $\pm$ 0.16 & 17.33 $\pm$ 0.10 & 7.92 $\pm$ 0.21 \\ \hline
\end{tabular}
}
\caption{Comparison of algorithm performance with or without explicit chemistry filters included in the reward signal as a scoring function. MolOpt is the normal benchmark and MolOpt-CF is the benchmark with chemistry filters and a diversity filter applied to each objective. Each algorithm was run 5 times with different seeds, the average score and variance was then summed over the 23 tasks to report the final score and standard deviation of the summed variance.}
\label{tab:molopt-CF}
\end{table*}

\subsection{Ablation study of the REINVENT algorithm}

One of the most popular RL algorithms used for the fine-tuning of CLMs is REINVENT \cite{olivecrona2017reinvent}. From a theoretical RL perspective, this algorithm can be viewed as a combination of the REINFORCE algorithm, experience replay, reward shaping and a sequence likelihood penalty term for the loss function. However, the impact of each specific mechanism is unclear. Therefore, using the modular components of ACEGEN, we conducted an ablation study to better understand the impact of reward shaping, experience replay, and sequence likelihood penalty, which are key components of the REINVENT algorithm. We used default REINVENT hyperparameters and measured performance on the MolOpt benchmark for comparison. Regarding performance, Table \ref{tab:molopt_ablation} and \autoref{fig:radars}d indicates that the primary improvement comes from adding experience replay to REINFORCE. While reward shaping does improve over REINFORCE, combining it with experience replay does not further improve performance. However, adding reward shaping that links the agent policy to the prior policy does improve the chemistry quality, measured by the proportion of valid and unique molecules passing the chemistry filters, albeit with a compromise in performance. These results highlight the importance of regularization in practical applications, as discussed in the previous section. In the original REINVENT implementation, the authors also include a loss term to penalize highly likely sequences. Table \ref{tab:molopt_ablation} shows that this primarily increases exploration but negatively affects all other metrics, including reducing the proportion of molecules passing chemistry filters. Furthermore, if used without reward shaping, it results in a complete collapse in learning.

\begin{table*}[ht!]
\centering
\small 
\resizebox{\linewidth}{!}{
\centering
\begin{tabular}{|l|c|c|c|c|c|c|c|}
\hline
 & REINFORCE & REINFORCE & REINFORCE & REINFORCE & REINFORCE & REINFORCE\footnotemark[1] \\
  & - & + replay & - & + replay & - & + replay \\ 
  & - & - & + shaping & + shaping & + shaping & + shaping \\ 
Metric  & - & - & - & - & + penalty & + penalty \\ 
\hline
Valid & 21.74 $\pm$ 0.04 & 21.77 $\pm$ 0.03 & \textbf{21.93 $\pm$ 0.02} & 21.90 $\pm$ 0.02 & 21.76 $\pm$ 0.02 & 21.78 $\pm$ 0.02 \\
Top-10 Avg & 14.74 $\pm$ 0.09 & \textbf{15.85 $\pm$ 0.10} & 14.80 $\pm$ 0.10 & 15.69 $\pm$ 0.10 & 14.79 $\pm$ 0.11 & 15.69 $\pm$ 0.11 \\
Top-10 AUC & 13.21 $\pm$ 0.07 & \textbf{13.67 $\pm$ 0.07} & 13.21 $\pm$ 0.07 & 13.57 $\pm$ 0.08 & 13.11 $\pm$ 0.07 & 13.60 $\pm$ 0.08 \\
Unique & 22.35 $\pm$ 0.05 & 22.28 $\pm$ 0.10 & 22.46 $\pm$ 0.05 & 22.29 $\pm$ 0.09 & \textbf{22.80 $\pm$ 0.05} & 22.63 $\pm$ 0.06 \\
\hline
B\&T-CF & 14.15 $\pm$ 0.23 & 14.34 $\pm$ 0.18 & \textbf{15.31 $\pm$ 0.15} & 15.08 $\pm$ 0.13 & 14.83 $\pm$ 0.13 & 14.74 $\pm$ 0.15 \\
B\&T-CF Top-10 Avg (Div) & 13.81 $\pm$ 0.09 & \textbf{14.95 $\pm$ 0.12} & 13.88 $\pm$ 0.09 & 14.80 $\pm$ 0.11 & 13.89 $\pm$ 0.12 & 14.83 $\pm$ 0.13 \\
B\&T-CF Top-10 AUC (Div) & 12.51 $\pm$ 0.07 & \textbf{12.91 $\pm$ 0.06} & 12.48 $\pm$ 0.06 & 12.83 $\pm$ 0.07 & 12.38 $\pm$ 0.07 & 12.85 $\pm$ 0.08 \\
B\&T-CF Diversity (SEDiv@1k) & 17.61 $\pm$ 0.18 & 17.39 $\pm$ 0.15 & 17.93 $\pm$ 0.10 & 17.52 $\pm$ 0.11 & \textbf{18.76 $\pm$ 0.11} & 18.19 $\pm$ 0.12 \\ \hline
\end{tabular}
}
\caption{Ablation study of REINFORCE algorithm on the MolOpt benchmark with and without separate components including experience replay, reward shaping, and a high sequence likelihood penalty. Each algorithm was run 5 times with different seeds, the average score and variance was then summed over the 23 tasks to report the final score and standard deviation of the summed variance. Note that all components together constitute the ACEGEN REINVENT implementation with default hyperparameters.}
\label{tab:molopt_ablation}
\end{table*}

\subsection{Case Study: De-novo generation in the 5-HT$_{2A}$}

To reflect more realistic situations that may arise in practical drug discovery scenarios, we apply the different RL algorithms implemented on a more challenging objective previously proposed by \cite{thomas2024molscore}. This objective is designing 5-HT$_{2A}$ receptor ligands selective over the highly related D$_2$ receptor utilizing only structural information. This is an important and relevant challenge as clinically, this off-target profile leads to undesirable extrapyramidal side-effects in the treatment of psychosis \cite{casey2022selective}. To achieve this, this objective aims to minimize the docking score against 5-HT$_{2A}$ (PDB: 6A93), yet maximize the docking score against D$_2$ (PDB: 6CM4), within a budget of 32,000 molecules. Both crystal structures are co-crystallized with Risperidone, reflecting the high degree of binding pocket similarity. In contrast to the proposed objective configuration, we use rDock \cite{ruiz2014rdock} due to more permissive licensing. We note that although docking scores are generally unreliable as a predictor of binding affinity, they still provide enrichment of selected molecules in virtual screening \cite{su2018comparative} and have shown benefits over the use of ligand-based prediction of binding affinity \cite{thomas2021comparison} which can lead to certain failure modes \cite{renz2019failure, langevin2022explaining}.

\begin{table*}[h]
\centering

\resizebox{\linewidth}{!}{

\begin{tabular}{|l|c|c|c|c|c|c|c|}
\hline
 & ACEGEN & ACEGEN & ACEGEN & ACEGEN & ACEGEN & ACEGEN & ACEGEN \\
Metric & REINFORCE & REINVENT & REINVENT-MolOpt & AHC & A2C & PPO & PPOD \\ 
\hline
Valid                & 0.95 & 0.96 & 0.93 & 0.94 & \textbf{0.98} & 0.94            & 0.95 \\
Top-10 Avg           & 1.05  & 1.09 & \textbf{1.10} & \textbf{1.10} & 1.04 & 1.06           & 1.07 \\
Top-10 AUC           & 1.02  & \textbf{1.05} & \textbf{1.05} & \textbf{1.05} & 1.02 & 1.01           & 1.02 \\
Unique               & \textbf{1.00} & \textbf{1.00} & 0.89 & \textbf{1.00} & \textbf{1.00} & 0.99            & 0.99 \\
\hline
B\&T-CF              & \textbf{0.71} & \textbf{0.71} & 0.45 & 0.63 & 0.27 & 0.59            & 0.48 \\
B\&T-CF Top-10 Avg (Div)     & 1.05  & 1.06 & 1.08 & \textbf{1.10} & 1.02 & 1.02 & 1.03 \\
B\&T-CF Top-10 AUC (Div)     & 1.01  & 1.03 & 1.04 & \textbf{1.05} & 1.00 & 0.99 & 0.99 \\
B\&T-CF Diversity (SEDiv@1k) & 0.89  & \textbf{0.91} & 0.64 & 0.89 & 0.77 & 0.60 & 0.53 \\ \hline
\end{tabular}
}
\caption{Performance summary of algorithms on 5-HT$_{2A}$ case study. Including after applying a basic chemistry filter (B-CF) as described previously. B-CF is the fraction of molecules that pass the chemistry filter. Note that the values here are re-normalized based on the scores achieved for a subset of 5-HT$_{2A}$ ligands that display at least 2-fold selectivity over D$_2$ as extracted from ChEMBL31 \cite{gaulton2012chembl}. Therefore, a score of 0.91 indicates the best score observed in the known ligand subset, and a score greater than 1.0 indicates that a single molecule achieves both a better (more negative) 5-HT$_{2A}$ docking score and (more positive) D$_2$ docking score than seen anywhere in the known subset.}
\label{tab:5HT2A_avg_metrics}
\end{table*}

\begin{figure*}
\centering
\includegraphics[width=0.75\textwidth]{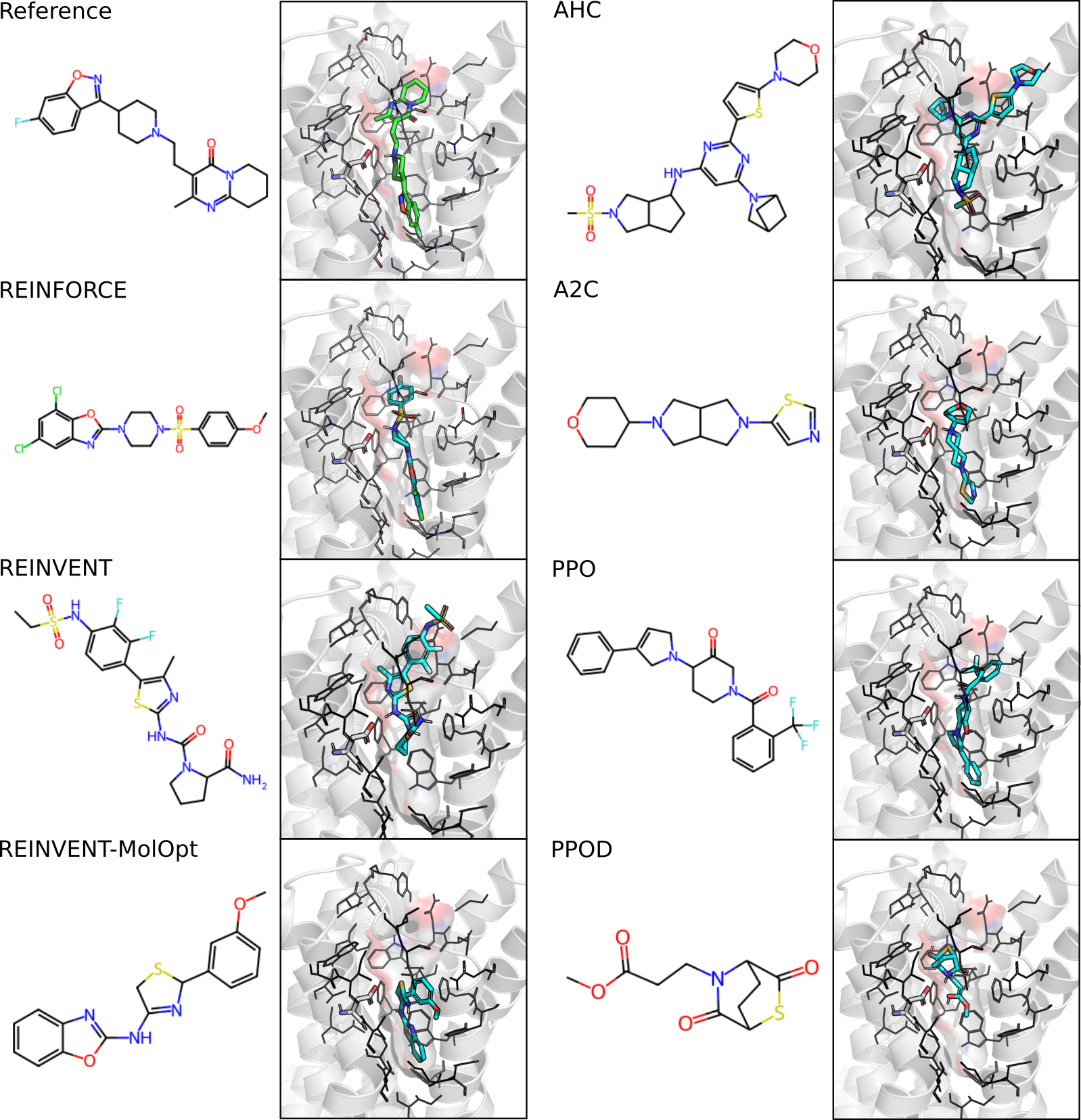}
\caption{Selected examples from the top 10 molecules on the 5HT$_{2A}$ selective task and their docked pose in 5-HT$_{2A}$ (PDB: 6A93). The co-crystallised ligand Risperidone is included as the reference.}\label{fig:5HT2A_POSES}
\end{figure*}

A summary of performance can be shown in the metrics in \autoref{tab:5HT2A_avg_metrics} and \autoref{fig:radars}c. These results show that when considering both optimization performance and chemistry - which is not explicitly captured in the objective - AHC achieves the highest average top 10 molecules and sample efficiency while maintaining high rates of molecules that pass chemistry filters and high diversity. This objective highlights the utility of compromising performance for regularization in practice, compared to the MolOpt benchmark which favors exploitation. A demonstration of \textit{de novo} molecules and their predicted docked pose in 5-HT$5_{2A}$ is shown in \autoref{fig:5HT2A_POSES}, for further methods, analysis and interpretation of the differences between RL algorithms from a chemistry perspective see section E in the supporting information.

\vspace{7mm}

\subsection{Case study: Scaffold constrained generation}
 ACEGEN can conduct constrained generation by leveraging PromptSMILES \cite{thomas2024promptsmiles} which proposes iterative CLM prompting on SMILES in combination with RL to achieve fine-tune a pre-trained CLMs to the task of constrained generation.
 Here we demonstrate this application with two experiments: 1) scaffold decoration via known synthetic reactions as proposed with LibINVENT \cite{fialkova2021libinvent}, and 2) scaffold decoration followed by scaffold-constrained docking to explore growth vectors inside a binding pocket. 

In the first experiment, we compared the performance of RL algorithms on two tasks proposed with LibINVENT \cite{fialkova2021libinvent}, both of which conduct scaffold decoration of a piperazine core with an amine linker moiety commonly present in D$_{2}$ receptor ligands. As in the MolOpt benchmark, we apply a budget of 10,000 molecules. The first task is to optimize the predicted probability of D$_2$ activity as evaluated by a QSAR model. The second task adds a selective reaction filter, rewarding piperazine growth via a Buchwald reaction and amine linker growth via an amide coupling reaction. Table \ref{tab:libinvent} shows successful optimization of both objectives, with AHC performing best on the first task, and REINVENT (either with default or MolOpt parameters) performing best on the second task. However, all algorithms except A2C are able to solve the task with a yield above 90\%, and fully satisfy the selective reaction objectives in 80 to 90\% of molecules.

\begin{table*}[h]
\centering
\resizebox{\linewidth}{!}{
\begin{tabular}{|ll|c|c|c|c|c|c|c|}
\hline
 & & ACEGEN & ACEGEN & ACEGEN & ACEGEN & ACEGEN & ACEGEN & ACEGEN \\
Task & Metric & REINFORCE & REINVENT & REINVENT-MolOpt & AHC & A2C & PPO & PPOD \\
\hline
\makecell[l]{D$_2$} &
Yield & 0.977 $\pm$ 0.005 & 0.989 $\pm$ 0.001 & 0.987 $\pm$ 0.007 & \textbf{0.991 $\pm$ 0.001} & 0.354 $\pm$ 0.125 & 0.942 $\pm$ 0.010 & 0.952 $\pm$ 0.010 \\ 
& Average score & 0.679 $\pm$ 0.008 & 0.720 $\pm$ 0.011 & 0.743 $\pm$ 0.020 & \textbf{0.794 $\pm$ 0.002} & 0.723 $\pm$ 0.043 & 0.749 $\pm$ 0.003 & 0.760 $\pm$ 0.006 \\ 
\hline
\makecell[l]{D$_2$ with}
& Yield & 0.972 $\pm$ 0.004 & \textbf{0.992 $\pm$ 0.001} & 0.988 $\pm$ 0.006 & 0.990 $\pm$ 0.003 & 0.555 $\pm$ 0.275 & 0.950 $\pm$ 0.015 & 0.947 $\pm$ 0.005 \\

reaction & Average score & 0.579 $\pm$ 0.003 & 0.668 $\pm$ 0.003 & \textbf{0.796 $\pm$ 0.007} & 0.723 $\pm$ 0.003 & 0.537 $\pm$ 0.149 & 0.705 $\pm$ 0.028 & 0.726 $\pm$ 0.013 \\

\makecell[l]{filters\\ \\} & \makecell[l]{Ratio of satisfied\\reaction filters} & 0.896 $\pm$ 0.005 & 0.753 $\pm$ 0.011 & \textbf{0.921 $\pm$ 0.009} & 0.809 $\pm$ 0.004 & 0.352 $\pm$ 0.196 & 0.830 $\pm$ 0.058 & 0.858 $\pm$ 0.020 \\ \hline
\end{tabular}
}
\caption{Algorithm comparison in combination with PromptSMILES for constrained molecule generation on LibINVENT DRD2 tasks (without and with selective reaction filters). Each algorithm was run 5 times with different seeds, the average value and standard deviation is reported.}
\label{tab:libinvent}
\end{table*}

\begin{figure*}
\includegraphics[width=\textwidth]{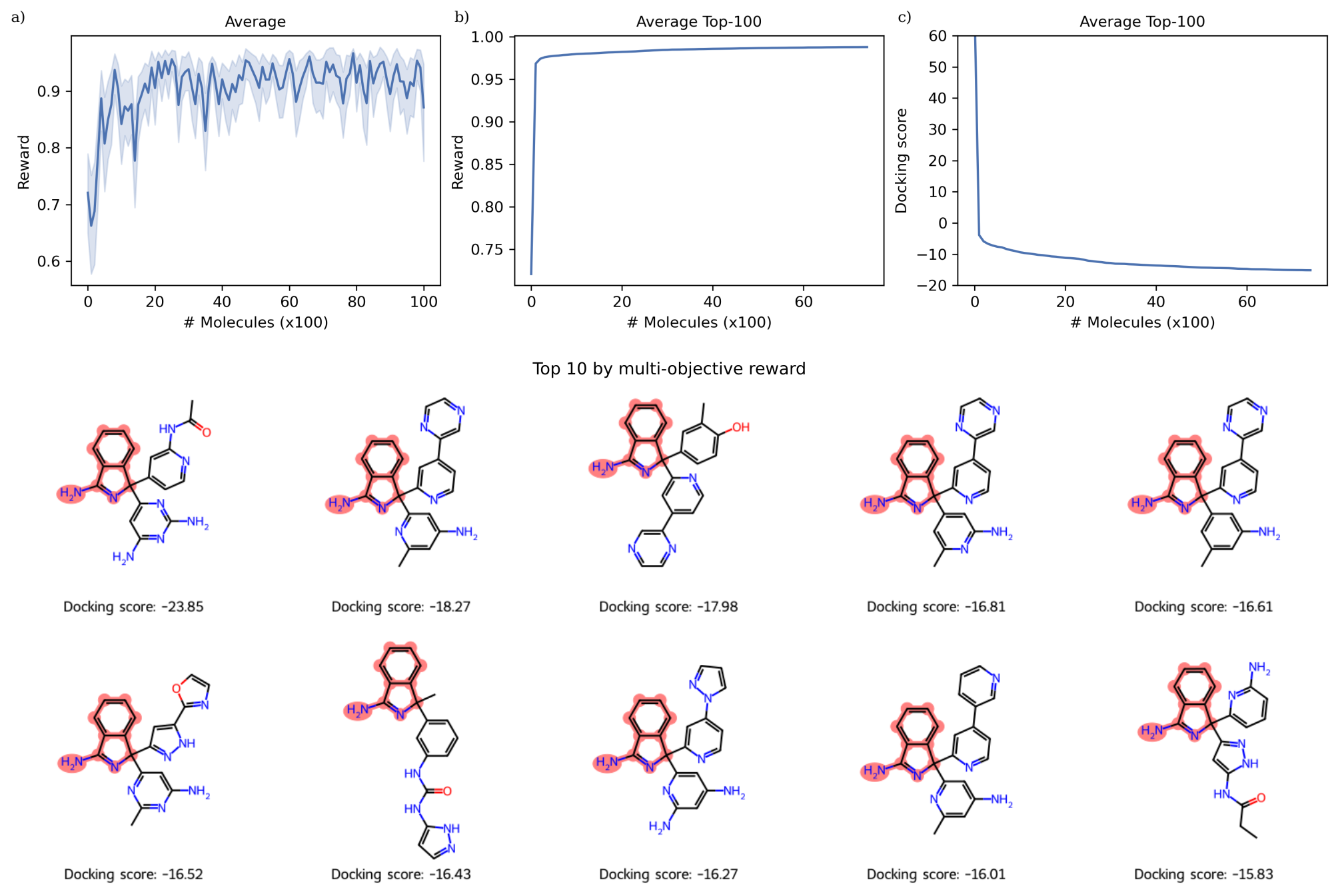}
\caption{Optimization of the multi-objective reward. The average reward and optimization of the underlying docking score are shown. The top 10 \textit{de novo} molecules are shown by multi-objective reward, with the constrained substructure highlighted in red and the docking score labeled below. For reference, the co-crystal ligand is re-docked with a docking score of -8.44. The docked poses are shown in Figure S13.}
\label{fig:BACE1}
\end{figure*}

In the second experiment, we used PromptSMILES in combination with AHC (based on the superior performance in the 5-HT$_{2A}$ case study) to conduct scaffold decoration in the catalytic site of BACE1, a key therapeutic target for the treatment of Alzheimer's disease. Here we used the reference structure 4B05 co-crystallized with AZD3839 \cite{jeppsson2012discovery} which is a typical BACE1 inhibitor consisting of an amidine core interacting with two catalytic aspartates, and occupation of the P2', P1 and P3 sub-pockets \cite{oehlrich2014evolution}. The bicyclic, amidine-containing core was used as a prompt for PromptSMILES, allowing two growth vectors, one into the P2' sub-pocket, and one into the P1 and P3 sub-pockets. Molecules were rewarded by minimizing the docking score using substructure constrained docking, as well as maintaining the number of heavy atoms and rotatable bonds in a sensible range (see supporting information F). Figure \ref{fig:BACE1} shows the successful maximization of the reward over the generation of 10,000 molecules (Figure \ref{fig:BACE1}a,b), and corresponding minimization of the docking score (Figure \ref{fig:BACE1}c). The top 10 molecules identified contain improved docking scores in the range -23.9 to -15.8 compared to the reference molecule re-docked with a docking score of -8.44. Docked poses (shown in Figure S13) show the recovery of aromatic rings in the P1 and P2 sub-pockets. This case study highlights the successful constrained optimization of a complex objective.

\section{Conclusion}

In this study, we introduce ACEGEN, a novel toolkit that combines the best methodologies of reinforcement learning (RL) within the field of drug discovery. Leveraging RL building blocks from TorchRL, a general highly-tested decision-making library, ACEGEN provides modular, efficient and versatile solutions for drug discovery that we demonstrate by implementing a suite of language-based solutions. We showcase ACEGEN's capabilities across diverse areas, including method benchmarking, algorithmic exploration, and real-world drug discovery applications. Our experiments contribute to a better understanding of popular algorithms like REINVENT, offer practical insights into selecting the most suitable algorithm for specific contexts, and underscore the importance of dependable and comprehensive toolkits. 

Overall, ACEGEN addresses various needs in drug discovery and effectively navigates the complex challenges associated with the field.
ACEGEN is a first step in the modularity required to explore the RL configuration space which we will utilize in future work to probe potential avenues for improvement. To achieve a performance improvement, it is also necessary to correctly measure the desired behavior which requires better benchmarks. For example, the best RL algorithms (REINVENT-MolOpt and PPOD) measured on the MolOpt benchmark are not the best as measured on the 5-HT$_{2A}$ docking benchmark highlighting a discrepancy in how we measure performance. Although we have introduced new chemistry-aware metrics in this work that better measure the real-world requirements of an RL algorithm in practice, new benchmarks are needed that better account for the exploration required as with the difficult 5-HT$_{2A}$ task, but without the computational expense. Lastly, ACEGEN currently only includes RL environments and architectures for CLMs which we are looking to extend to various other architectures in the future, in particular, it is straightforward to have generative models in 3D space.

\section{Data and Software Availability}

All software used in this manuscript is freely available open-source under an MIT license. ACEGEN is available at \url{https://github.com/Acellera/acegen-open}.  The parameters of the pre-trained model used in this work are available in ACEGEN repository. All resulting de novo molecules from the experiments are freely accessible on Zenodo
\href{https://zenodo.org/records/11243056?token=eyJhbGciOiJIUzUxMiJ9.eyJpZCI6IjZkNzQ2MjE1LTljNjEtNDE5Ni1hZjU3LTI0MWYzZWY5YWY4OSIsImRhdGEiOnt9LCJyYW5kb20iOiI1NDI1MWQ2ZWNhNmRlOWQzMDhlZTU0ZTg1MWViZmJhMCJ9.rwZOGe7OdKJvGZSvo-0HHHvfMTi95XRPKA9Vp4dNJ7wBMSb_z0hd606tXMPoZnWAmRwKbP_LD4-ZVRv6tZ3hKw}{here}. The full results of the MolOpt benchmark for all computed metrics and seeds are available in \textbf{Supporting Data 1}.

\section{Supporting information}
The supporting information file includes additional experimental details, results, and code examples. It also contains supplementary figures referenced in the text.

\section{Funding}
M.T. is partly funded by the Flanders innovation \& entrepreneurship (VLAIO) project HBC.2021.112.

\clearpage

\bibliography{main}

\end{document}